\documentclass{article}

\PassOptionsToPackage{numbers,sort&compress}{natbib}

\usepackage[final,main]{neurips_2026}

\usepackage[utf8]{inputenc} 
\usepackage[T1]{fontenc}    
\usepackage{hyperref}       
\usepackage{url}            
\usepackage{booktabs}       
\usepackage{amsfonts}       
\usepackage{nicefrac}       
\usepackage{microtype}      
\usepackage[table]{xcolor}  
\usepackage{enumitem}
\usepackage{amsmath}
\usepackage{amssymb}
\usepackage{graphicx}
\usepackage{multirow}
\usepackage{colortbl}
\usepackage{caption}
\usepackage{pifont}%
\title{Masked Diffusion Vision-Language Models \\ for Temporal Action Localization}

\author{Fengshun Wang$^{1}$\thanks{Equal contribution.} , Zhengbo Zhang$^{2}$\footnotemark[1] , Zhigang Tu$^{1}$\thanks{Corresponding Author.} \\
$^1$Wuhan University~$^2$Singapore University of Technology and Design~ \\
}

\begin{document}

\maketitle

\begin{center}
\vspace{-6mm}
    \centering
    \includegraphics[width=\linewidth]{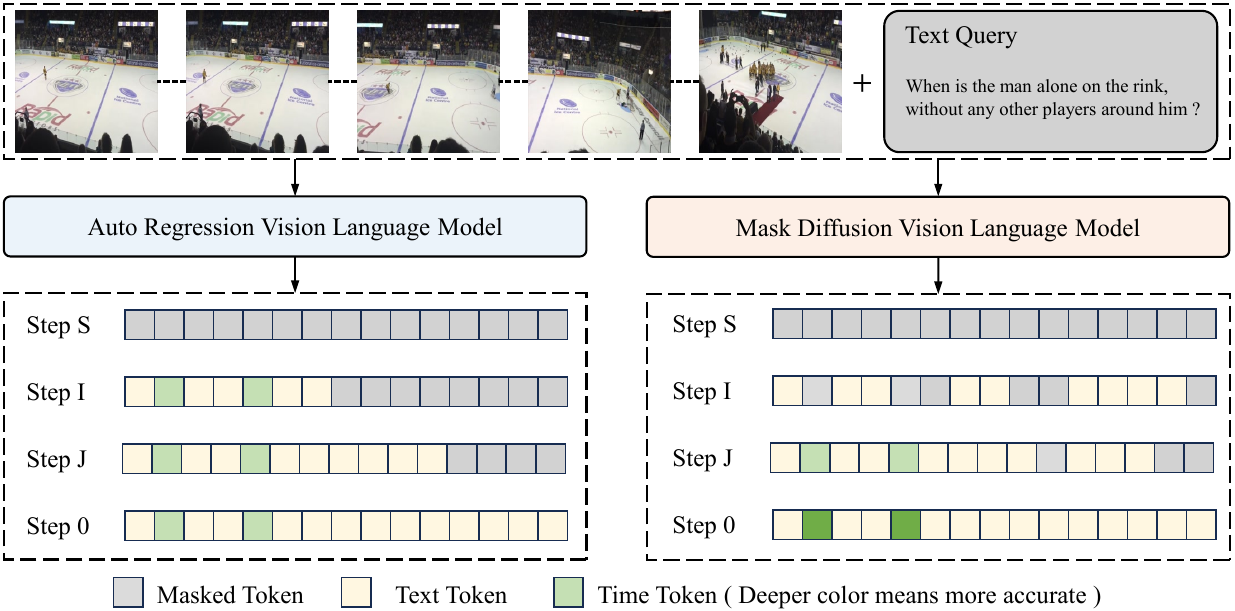}
    \captionof{figure}{Autoregressive vision-language approaches generate tokens from left to right, limiting their ability to refine earlier predictions using future context. In contrast, masked diffusion vision-language models attend to the full sequence at every step with bidirectional attention, allowing joint refinement of temporal boundaries and semantic descriptions.}
    \label{fig:teaser}
\end{center}
\begin{abstract}
  Temporal action localization (TAL) requires recognizing the target event and localizing its start and end times precisely in untrimmed videos. Recent vision-language formulations improve semantic reasoning and support language-conditioned outputs, but their autoregressive decoders still generate tokens from left to right, preventing later semantic evidence from revising earlier timestamp predictions.
  We adapt masked diffusion vision-language models (MDVLMs) to TAL so that semantic tokens and boundary tokens remain editable throughout iterative denoising with bidirectional attention, allowing temporal boundaries and semantic content to be refined jointly.
  Direct adaptation, however, creates two TAL-specific mismatches: standard masked diffusion training corrupts all positions uniformly at random, but the time tokens are more reliable when enough semantic context is available; and token-level cross-entropy does not reflect temporal IoU. To address these mismatches, we introduce a \textbf{Planned Training Objective} that uses boundary-aware masking and step-weighted reconstruction to rehearse the late recovery of time tokens, together with a \textbf{Step-Level IoU Reward} that provides overlap-aware supervision during denoising. A standard sequence-level cross-entropy term provides the base reconstruction signal.
  Experiments on ActivityNet-RTL, ActivityNet-1.3, and THUMOS-14 show that MDVLM-TAL improves both temporal reasoning and boundary localization over autoregressive vision-language baselines, with especially strong gains under stricter temporal IoU criteria. Our source code is available at \url{https://github.com/ycwfs/MDVLM-TAL}.
\end{abstract}

\section{Introduction}

Temporal Action Localization (TAL) requires localizing the temporal extent of each action instance in an untrimmed video. A correct prediction couples two decisions, recognizing the queried event category and placing its start and end times precisely. Because boundary errors of even a few seconds push the temporal IoU below standard evaluation thresholds, TAL demands stricter temporal precision than video-level classification or captioning.

Classical TAL pipelines treat localization as visual proposal classification and regression~\cite{chao2018rethinking,lin2019bmn,xu2020g,Zhang2022_E4DJG4I4}. They extract temporal features, generate candidate segments or regress boundary offsets, and classify each segment with action-specific heads. These detectors excel on closed-set benchmarks. However, they are not well-suited for settings that require language conditioned undertanding, detailed event semantics.
To deal with the problems above, some researchers have innovated using the powerful contextual understanding ability of Large Vision-Language Models (LVLM) by treating TAL as generative task with LVLM~\cite{Huang2024_RP67QJDE,Zeng2024_W77Z2KFV,Guo2024_ZZVDVZ6M,Huang2024_E99RJNWL,Zhang2023_VideoLLaMA,Li2023_VideoChat,Yu2023_SeViLA,Xu2024_I9Z8BZ8F,Xu2024_FM5VVXFF,Liu2023_LLaVA} (see Figure \ref {fig:teaser}). The video and query are encoded jointly, timestamps are discretized into special time tokens, and the action label, boundary tokens, and optional text are generated as one sequence in~\cite{Huang2024_RP67QJDE}. This formulation grants TAL access to language priors and free form explanations. However, LVLM-based TAL systems still inherit the autoregressive (AR) decoder of their backbone, leaving a gap between the context understanding ability and the temporal action localization ability of the AR model~\cite{wang2022internvideo,wang2024internvideo2,wang2025internvideo2,bai2025qwen3}. This gap is caused by the left to right token generation order of AR and the immutability of token generation. And it can be optimized through iterative updates.

As shown in Figure \ref {fig:teaser}, the recent language model paradigm, Masked diffusion vision-language models (MDVLMs)~\cite{li2025lavida}, can replace this left to right commitment with iterative denoising under bidirectional attention~\cite{nie2025lldm}. Generation begins from a fully masked response and progressively unmasks tokens with global context at every step, so semantic tokens and time tokens remain editable until the final step. This is the structural property TAL requires, boundary tokens are revised after more semantic evidence has appeared, and the start, end, and text tokens are refined within the same generation process.

With these in mind, in this paper, we aim to leverage MDVLMs to address TAL. Yet, this is non-trivial, treating TAL as a generic masked diffusion problem leaves two TAL-specific mismatches unaddressed. \textbf{(i) Schedule mismatch.} The response is a \emph{mixed} sequence of ordinary text tokens and time tokens that encode boundaries. Our dataset-level statistics show that semantic tokens are revealed earlier than time tokens and that boundary predictions stabilize later in denoising (Figure~\ref{fig:dataset_recovery}), whereas standard masked diffusion training corrupts all positions uniformly at random and never rehearses this late reveal behavior. \textbf{(ii) Loss mismatch.} TAL is evaluated by temporal IoU, which depends on the numerical \emph{distance} between predicted and ground-truth boundaries, but token-level losses penalize all wrong tokens equally and fail to reward an update from $\langle t_{0}\rangle$ to $\langle t_{47}\rangle$ when the target is $\langle t_{43}\rangle$.

To tackle these challenges, we propose \textbf{MDVLM-TAL}, a masked diffusion vision-language framework that closes both gaps in the training objective. The \textbf{Planned Training Objective} uses boundary-aware masking and step-weighted reconstruction so that time tokens are typically recovered later than semantic tokens during training, matching their behavior at inference. The \textbf{Step-Level IoU Reward} provides overlap-aware supervision during denoising. A standard sequence-level cross-entropy term provides the base reconstruction loss. Our contributions are:
\begin{itemize}[leftmargin=*,noitemsep,topsep=2pt]
    \item We formulate TAL as masked diffusion sequence generation, so semantic tokens and boundary time tokens are refined jointly through bidirectional denoising rather than fixed by a left to right order.
    \item We introduce a \textbf{Planned Training Objective} that uses boundary-aware masking and step-weighted reconstruction to delay the recovery of time tokens relative to semantic tokens.
    \item We propose a \textbf{Step-Level IoU Reward} that injects overlap-aware supervision into intermediate denoising states for more precise temporal localization.
    \item On THUMOS-14, ActivityNet-1.3, and ActivityNet-RTL, MDVLM-TAL surpasses both autoregressive vision-language baselines and the strongest specialized TAL detectors, with the largest gains at strict tIoU thresholds.
\end{itemize}

\section{Related Work}

\paragraph{Temporal Action Localization.}
Temporal action localization has evolved from proposal-based and two-stage pipelines, such as TAL-Net~\cite{chao2018rethinking}, BMN~\cite{lin2019bmn}, G-TAD~\cite{xu2020g}, RTD-Net~\cite{tan2021relaxed}, TCANet~\cite{qing2021temporal}, and ContextLoc~\cite{zhu2021enriching}, to stronger one-stage and transformer-centric detectors, including ReAct~\cite{shi2022react}, ActionFormer~\cite{Zhang2022_E4DJG4I4}, TALLFormer~\cite{Tallformer}, TadTR~\cite{FX979ZEP}, TriDet~\cite{Tridet}, TemporalMaxer~\cite{TemporalMaxer}, TFFormer~\cite{TFFormer}, TransGMC~\cite{TransGMC}, RefineTAD~\cite{H8Q99P63}, DualDETR~\cite{U84863CG}, Pred-DETR~\cite{CV922R8W}, and CLTDR~\cite{Li2024_CLTDR}. Recent work further expands TAL beyond the standard fully supervised closed-set setting, studying weakly supervised localization~\cite{UBV766ZB,9NG3983D,XTT9IIB7,ZEUUYWHT,DTVJ7N9S}, zero-shot or open-vocabulary detection~\cite{JX7MI83H,37QXWBMM,6SZNS42I}, online localization~\cite{JXFQ498U,5F2AFWKI}, and generative or denoising formulations such as DiffTAD and DenoiseLoc~\cite{Nag2023_DiffTAD,7QIS3HW9}. OpenTAD~\cite{V796WXUR} further shows that evaluation conclusions can change substantially once methods are reimplemented under a unified framework. Still, most TAL methods remain detector-style models that directly classify and regress intervals from visual features, rather than generating temporally grounded multimodal outputs.

\paragraph{Vision-Language Models for Temporal Action Localization.}
Recent generative vision-language methods represent boundaries as timestamp tokens and cast localization as text generation, including LITA, VTimeLLM, VTG-LLM, and UniMD~\cite{Huang2024_RP67QJDE,Huang2024_E99RJNWL,Guo2024_ZZVDVZ6M,Zeng2024_W77Z2KFV}. A related line uses MLLMs as auxiliary supervision for weakly supervised TAL rather than as the decoder itself~\cite{Zhang2024_XRTWDU53}. These methods show that vision-language modeling is useful for temporal localization, but they mostly rely on autoregressive vision-language decoding or use the vision-language model as an external guide. Our work is closest to the generative line, but replaces left to right decoding with masked diffusion.

\paragraph{Diffusion Models for Perception and Structured Generation.}
Text-to-image diffusion models also provide useful representations for visual perception. VPD~\cite{zhao2023unleashing} adapts their pretrained denoising features and cross-attention maps to perception tasks, while TADP~\cite{kondapaneni2024text} improves these representations through caption-based text--image alignment. Related work applies text-to-image diffusion models to category-agnostic pose estimation~\cite{peng2024harnessing} and unsupervised visual tracking, where Diff-Tracker learns and updates target-specific prompts~\cite{zhang2024diff}. These studies demonstrate the value of generative semantic knowledge for spatial prediction. For human motion generation, DC-Motion~\cite{wang2026dc} separates discrete semantic structure from continuous motion details, using masked modeling for structural tokens and diffusion for residual dynamics. This distinction between semantics and fine-grained details is related to our separation of semantic and time tokens. MDVLM-TAL addresses temporal localization by jointly denoising both token types within a language response, with a task-specific recovery schedule and temporal-overlap supervision.

\begin{figure}
    \centering
    \includegraphics[width=\linewidth]{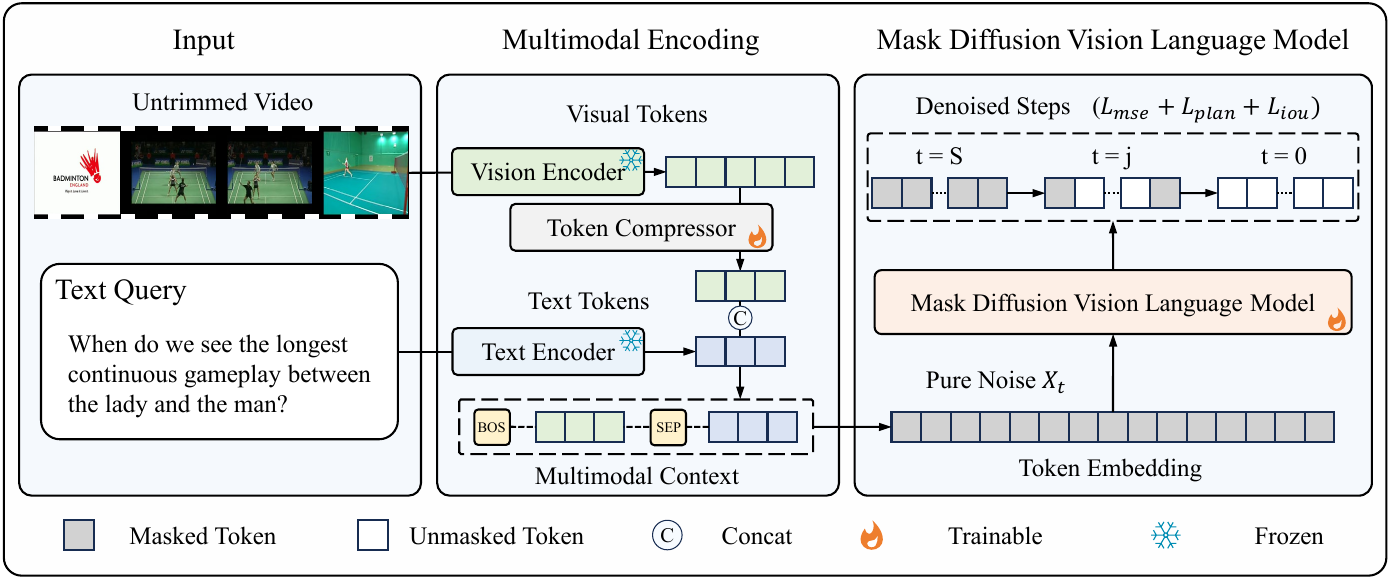}
    \caption{Overview of MDVLM-TAL. Given a video and a query, a vision and text encoder are used to encode the inputs. Then, we compress the vision token with a token compressor. Video and text tokens are concatenated with noisy tokens to build a multimodal context. The masked diffusion backbone jointly denoises semantic tokens and boundary time tokens using global bidirectional attention. The Planned Training Objective keeps time tokens masked longer than semantic tokens during training, and the Step-Level IoU Reward provides overlap-aware supervision during denoising.}
    \label{fig:architecture}
\end{figure}

\section{Method}
\label{sec:method}

\subsection{Preliminaries}
\noindent \textbf{Task definition.} Given an untrimmed video $V=\{f_n\}_{n=1}^{T_v}$ and an instruction $q$, our goal is to generate a response that contains temporal boundaries and optional explanation text. We consider two regimes. In reasoning temporal localization (e.g., ActivityNet-RTL), $q$ specifies one target event and the response contains one boundary pair plus text. In standard closed-set TAL benchmarks, $q$ is class-agnostic and the model must infer all action instances directly from video content.

\noindent \textbf{Time-token formulation.} We discretize the timeline into $N$ bins and represent boundaries with special tokens from $\mathcal{V}_{\text{time}}=\{\langle t_0\rangle,\dots,\langle t_{N-1}\rangle\}$. Let $\mathcal{V}$ denote the full output vocabulary, let $Y^\star = [y_1,\dots,y_{L}]$ denote the target response sequence, and let $\hat{B}=[\hat{b}_s,\hat{b}_e]$ be the segment decoded from the generated time tokens. If the ground-truth boundary indices are $(s^\star,e^\star)$, the normalized timestamps are
$
b_s^\star=\frac{s^\star}{N-1}, b_e^\star=\frac{e^\star}{N-1}, 0\le b_s^\star\le b_e^\star\le 1.
$
Localization quality is measured by the temporal IoU between the predicted segment $\hat{B}$ and the ground-truth segment $B^\star=[b_s^\star,b_e^\star]$.

\noindent \textbf{Masked Diffusion Large Language Models.} Let $x_0=Y^\star$ be the clean target response. A masked diffusion language model defines a sequence of corrupted states $\{x_t\}_{t=0}^{S}$, where $t=0$ is the clean sequence and $t=S$ is the fully masked state. For each diffusion step $t\in\{1,\dots,S\}$, we sample a binary keep mask $m_t\in\{0,1\}^{L}$ and construct
\begin{equation}
x_t = m_t \odot x_0 + (1-m_t)\odot \texttt{[MASK]}.
\end{equation}
We denote the masked positions by $\mathcal{M}_t=\{i\in\{1,\dots,L\}:m_{t,i}=0\}$. Conditioned on the multimodal context $C$ and the current step $t$, the denoiser parameterized by $\theta$ predicts the clean-token distribution
\begin{equation}
p_\theta(x_0\mid x_t,C,t)=\prod_{i=1}^{L}p_\theta(x_{0,i}\mid x_t,C,t).
\end{equation}
Training samples a diffusion step and learns to recover the clean tokens at the masked positions. Inference starts from the fully masked state $x_S$ and runs reverse denoising from $t=S$ to $t=1$, progressively revealing the masked positions with the highest token confidence until the final sequence is obtained. Because every position remains editable until the last steps, masked diffusion is well suited to TAL, where semantic tokens and boundary time tokens should be refined jointly rather than fixed by a left-to-right decoding order.

\subsection{Architecture overview.}
As shown in the Figure\ref{fig:architecture}, we follow the standard \emph{visual encoder + multimodal projector + language backbone} design used in Video-LLM systems~\cite{Liu2023_LLaVA,Zhang2023_VideoLLaMA,Xu2024_I9Z8BZ8F}, we encode the video as $Z_v = \Pi\!\left(\operatorname{Enc}_v(V)\right)$, and a token compressor is used to compress and project vision tokens, then we tokenize the instruction as $Z_q$, and form the multimodal context $C=[\texttt{BOS}, Z_v, \texttt{SEP}, Z_q, \texttt{SEP}]$, where \texttt{BOS} and \texttt{SEP} denote the beginning-of-sequence and separator tokens.

The token compressor consists of one cross-attention block applied independently to each frame. Eight learned query tokens, shared across frames, attend to the spatial tokens of each frame. A feed-forward layer with a residual connection follows the cross-attention operation. This produces eight visual tokens per frame, which we concatenate in frame order to preserve the temporal ordering of the video.

\subsection{Planned Training Objective}
Standard masked diffusion training uses a uniform random masking process, but in TAL boundary tokens usually become reliable only after enough semantic context has been formed. If timestamps appear too early during training, the learned corruption path no longer matches the late-reveal behavior of boundary tokens during reverse denoising. Our planned training objective addresses this mismatch by delaying the unmasking of boundary tokens and emphasizing denoising stages where fine temporal cues matter most.

Planned training has two parts: a boundary-aware masking policy and a step-weighted reconstruction loss. For a sampled denoising step $t$, the masking policy keeps temporal boundary positions masked more aggressively than non-temporal-boundary positions, and the reconstruction loss places larger weight on lower-noise steps that carry finer temporal boundary information.

\noindent \textbf{Boundary-aware masking.} We partition token positions into a temporal boundary set
$
\mathcal{B}=\{i\in\{1,\dots,L\}:y_i^\star\in\mathcal{V}_{\text{time}}\},
$
whose target token is a time token, and its complement $\mathcal{N}=\{1,\dots,L\}\setminus\mathcal{B}$. For step $t$, we define type-specific keep probabilities as
$
\kappa_t^{\mathcal{B}}=\left(1-\frac{t}{S}\right)^{\gamma},
\kappa_t^{\mathcal{N}}=\left(1-\frac{t}{S}\right)^{\eta}, \gamma>\eta>0,
$
so $\kappa_t^{\mathcal{B}}<\kappa_t^{\mathcal{N}}$ for $t\in(0,S)$ and boundary tokens remain masked longer than non-boundary tokens. We then sample the training mask as
\begin{equation}
m_{t,i}\sim
\begin{cases}
\operatorname{Bernoulli}(\kappa_t^{\mathcal{B}}), & i\in\mathcal{B},\\
\operatorname{Bernoulli}(\kappa_t^{\mathcal{N}}), & i\in\mathcal{N}.
\end{cases}
\end{equation}
This induces a boundary-aware corruption distribution
$
q_{\phi}(x_t \mid x_0,t)=\prod_{i=1}^{L} q_{\phi}(x_{t,i}\mid x_{0,i},t),
$
where $\phi=\{\gamma,\eta\}$ parameterizes the planned masking policy and $x_t$ is formed from $m_t$ as in the preliminaries.

\noindent \textbf{Step-weighted reconstruction.} We train the denoiser under the planned corruption process with cross-entropy restricted to the masked positions:
\begin{equation}
\mathcal{L}_{\text{plan}}
=
\mathbb{E}_{(V,q,Y^\star)\sim\mathcal{D}}
\mathbb{E}_{t\sim\mathcal{U}\{1,\dots,S\}}
\mathbb{E}_{x_t\sim q_{\phi}(\cdot\mid x_0,t)}
\left[
w_t \!\!\sum_{i:\,m_{t,i}=0}\!\!-\log p_\theta(x_{0,i}\mid x_t,C,t)
\right],
\end{equation}
where $\sum_{t=1}^{S}w_t=1$ and $w_{t_1}>w_{t_2}$ if $t_1<t_2$, so lower-noise denoising stages receive larger weight. Compared with vanilla diffusion training, $\mathcal{L}_{\text{plan}}$ makes the model rehearse the late recovery of time tokens observed during reverse denoising, instead of treating time tokens and semantic tokens as equally easy to reveal at every step.

\subsection{Step-Level IoU Reward}
Cross-entropy supervises token identity rather than temporal distance. In TAL, predicting $\langle t_{47}\rangle$ is much better than predicting $\langle t_{0}\rangle$ when ground truth is $\langle t_{43}\rangle$, but standard token losses do not capture this difference. We therefore add step-level supervision that measures whether each denoising update improves the current temporal segment.

Our IoU-guided reward has two parts: a soft IoU score derived from the start and end token distributions at each denoising step, and a gating mechanism that activates the reward only after both boundary tokens have been revealed.

\noindent \textbf{Soft boundary reward.} Let $i_s$ and $i_e$ denote the start and end slots of the current action tuple in the response template. From the full-vocabulary logits $z_t^{(i_s)}$ and $z_t^{(i_e)}$, we restrict to the time-token vocabulary $\{\langle t_k\rangle\}_{k=0}^{N-1}$ and renormalize:
\begin{equation}
p_t^s(k)=\frac{\exp z_t^{(i_s)}[\langle t_k\rangle]}{\sum_{k'=0}^{N-1}\exp z_t^{(i_s)}[\langle t_{k'}\rangle]},
\qquad
p_t^e(k)=\frac{\exp z_t^{(i_e)}[\langle t_k\rangle]}{\sum_{k'=0}^{N-1}\exp z_t^{(i_e)}[\langle t_{k'}\rangle]}.
\end{equation}
\noindent We then compute ordered soft boundaries by expectation as $\tilde{s}_t=\sum_{k=0}^{N-1}k\,p_t^s(k)$, $\tilde{e}_t=\sum_{k=0}^{N-1}k\,p_t^e(k)$, $\bar{b}_{s,t}=\frac{\tilde{s}_t}{N-1}$, and $\bar{b}_{e,t}=\frac{\tilde{e}_t}{N-1}$. We keep the start and end roles distinct and define the soft overlap terms as
\begin{align}
\ell_t^{\text{pred}}=\max\!\left(0,\bar{b}_{e,t}-\bar{b}_{s,t}\right),
\ell_t^{\cap}&=\max\!\left(0,\min\!\left(\bar{b}_{e,t},b_e^\star\right)-\max\!\left(\bar{b}_{s,t},b_s^\star\right)\right),\nonumber \\
\ell_t^{\cup}&=\ell_t^{\text{pred}}+\left(b_e^\star-b_s^\star\right)-\ell_t^{\cap},
\end{align}
and compute the soft IoU reward as
$
r_t^{\text{soft}}=\frac{\ell_t^{\cap}}{\ell_t^{\cup}+\epsilon}.
$

\noindent \textbf{Gated relative weighting.} Let $\hat{y}_t^{(s)}$ and $\hat{y}_t^{(e)}$ denote the currently decoded tokens at the start and end positions after reverse-denoising step $t$. To avoid noisy supervision from incomplete boundaries, we activate IoU reward only when both boundary tokens are generated at step $t$. Let $g_t\in\{0,1\}$ be
$
g_t=\mathbb{I}\!\left[\hat{y}_t^{(s)}\neq \texttt{[MASK]} \;\land\; \hat{y}_t^{(e)}\neq \texttt{[MASK]}\right].
$
Next, we compute the group-relative reward statistics over the active denoising steps:
\begin{equation}
\bar{r}=\frac{\sum_{t=1}^{S} g_t r_t^{\text{soft}}}{\epsilon+\sum_{t=1}^{S} g_t},
\qquad
\sigma_r=\sqrt{\frac{\sum_{t=1}^{S} g_t \left(r_t^{\text{soft}}-\bar{r}\right)^2}{\epsilon+\sum_{t=1}^{S} g_t}+\epsilon},
\end{equation}
and define the normalized step advantage $A_t = g_t\,\frac{r_t^{\text{soft}}-\bar{r}}{\sigma_r}$. To keep the relative weighting strictly non-negative, we use $\tilde{A}_t=\max(A_t,0)$.
We optimize a unified IoU reward objective:
\begin{align}
\mathcal{L}_{\text{iou}}
&=
\frac{1}{\epsilon+\sum_{t=1}^{S}g_t}
\Bigg[
\sum_{t=1}^{S}\alpha_t g_t(1-r_t^{\text{soft}}) +\mu\sum_{t=2}^{S}g_t g_{t-1}\max\!\left(0,\,r_t^{\text{soft}}-r_{t-1}^{\text{soft}}+\delta\right) \nonumber\\
&\quad -\lambda_{\text{rel}}\sum_{t=1}^{S}\mathrm{stopgrad}(\tilde{A}_t)\log\!\big(p_t^s(s^\star)\,p_t^e(e^\star)\big)
\Bigg],
\end{align}
\noindent Here $\alpha_t$ are step weights, $\mu$ controls the monotonic refinement penalty, $\delta$ is the refinement margin, and $\lambda_{\text{rel}}$ controls the relative weighting term. The first term favors high step-wise soft tIoU at every active step. The second encourages soft IoU to improve along the reverse trajectory (from $x_t$ to $x_{t-1}$, i.e., toward the clean state at $t=0$): the hinge $\max(0,r_t^{\text{soft}}-r_{t-1}^{\text{soft}}+\delta)$ is non-zero whenever $r_{t-1}^{\text{soft}}<r_t^{\text{soft}}+\delta$, penalizing pairs where overlap fails to improve by at least $\delta$. The third reinforces active steps whose IoU is above the active-step mean without pushing weak steps away from the target. Here $\mathrm{stopgrad}(\cdot)$ means that the normalized weights are detached before they are applied to the boundary log-likelihood.

\subsection{Final objective.}
Besides the planned training term, the implementation keeps the standard token-level reconstruction loss returned by the backbone. Let $\bar{Y}_t^\star$ denote the supervised label sequence after ignoring positions outside the current training target. We write the base reconstruction loss as
$
\mathcal{L}_{\text{ce}}
=
\operatorname{CE}\!\left(p_\theta(\cdot\mid x_t,C,t), \bar{Y}_t^\star\right).
$
And the full training objective is
$
\mathcal{L}
=
\lambda_{\text{ce}}\mathcal{L}_{\text{ce}}
+\lambda_{\text{plan}}\mathcal{L}_{\text{plan}}
+\lambda_{\text{iou}}\mathcal{L}_{\text{iou}}.
$

\subsection{Inference}
Given an input video $V$ and instruction $q$, we first build the multimodal context $C$ from the encoded video tokens and instruction tokens. Inference starts from the fully masked output sequence $x_S$ and runs reverse denoising for $t=S,S-1,\dots,1$. At each step, the denoiser predicts $p_\theta(x_{0,i}\mid x_t,C,t)$ for every masked position. We then reveal the scheduled number of masked positions with highest maximum token probability and keep the remaining positions masked to form $x_{t-1}$. This process usually resolves semantic tokens earlier and time tokens later, which matches the planned training design. After the final step, the decoded $x_0$ is parsed into the output response, including the predicted boundary tokens and any associated text.

\section{Experiments}
\subsection{Implementation Details}
\noindent \textbf{Datasets and Benchmarks.}

We evaluate MDVLM-TAL on three TAL benchmarks: ActivityNet-1.3, THUMOS-14, and ActivityNet-RTL. ActivityNet-1.3 is a large-scale benchmark built on the ActivityNet video collection~\cite{Heilbron2015_CX49Q6U2} and emphasizes long untrimmed videos with diverse action classes. THUMOS-14 contains short, densely annotated sports actions and stresses boundary precision under frequent action transitions. ActivityNet-RTL~\cite{Huang2024_RP67QJDE} focuses on reasoning and evaluates whether models can jointly produce semantic rationales and accurate temporal boundaries.

\noindent \textbf{Evaluation Metrics.}
For THUMOS-14 and ActivityNet-1.3 we follow the standard TAL protocol and report per-class mean Average Precision (mAP) at multiple tIoU thresholds, with the official averages over $[0.3{:}0.1{:}0.7]$ and $[0.5{:}0.05{:}0.95]$ respectively. For ActivityNet-RTL we follow the LITA protocol~\cite{Huang2024_RP67QJDE} and report mIoU, $P@0.5$, and the GPT-based ``Score'' for explanation quality under per-video averaging; we additionally report $P@0.75$ and $P@0.95$ for our model and mark them ``--'' for baselines that did not.

\noindent \textbf{Settings.}
We use \texttt{LLaDA-8B-Instruct}~\cite{nie2025lldm} as the masked diffusion language backbone and \texttt{SigLIP-SO400M-Patch14-384}~\cite{zhai2023siglip} as the vision encoder. In our implementation, $\theta$ in the the denoiser consists of the LoRA adapters inserted into the language model, together with the multimodal projector and the temporal video compressor when it is enabled. The pretrained backbone weights and the SigLIP vision-tower weights remain frozen by default.

We uniformly sample 100 frames from each video, discretize timestamps into $N=100$ relative time tokens $\langle t_0\rangle,\dots,\langle t_{99}\rangle$, use $S=64$ denoising steps, and set the number of compressor query tokens to 8. For a timestamp $\tau$ in a video of duration $D$, we use $k=\mathrm{round}((N-1)\tau/D)$ during encoding and decode a predicted token by $\tau \approx Dk/(N-1)$. We use LoRA~\cite{hu2022lora} with rank 16, $\alpha=32$, and dropout 0.05. For the planned masking policy, we fix $(\gamma,\eta)=(2,1)$, and for the step-weight schedule we use the normalized linear form $w_t=\frac{2(S-t+1)}{S(S+1)}.$
For the IoU reward, we use the same step weights $\alpha_t=w_t$, together with $\mu=0.5$, $\delta=0.01$, $\lambda_{\text{rel}}=0.5$, and $\epsilon=10^{-8}$. Unless otherwise stated, the loss-balance coefficients are $(\lambda_{\text{plan}},\lambda_{\text{iou}},\lambda_{\text{ce}})=(1.0,0.5,0.2)$.

For closed-set TAL, the prompt is ``Identify all action instances in the video and output each action label with its start and end time stamps.'' Across closed-set TAL and ActivityNet-RTL, we use AdamW with learning rate $2\times 10^{-4}$, weight decay $0.01$, cosine decay, warmup ratio $0.03$, per-device batch size 1, and gradient accumulation of 8 steps. In the released scripts, RTL uses 5 epochs with maximum sequence length 4096, while the closed-set event-localization script uses 10 epochs with maximum sequence length 2048. For ablations over loss balance, we report coefficients $(\lambda_{\text{plan}},\lambda_{\text{iou}},\lambda_{\text{ce}})$ in Table~\ref{tab:ablation_weights}. For RTL inference, we use confidence-based unmask decoding with block length 32 and step ratio 1. All the experiments are conducted on 4 NVIDIA L40S GPUs.

\begin{table*}[t]
\centering
\scriptsize
\setlength{\tabcolsep}{3pt}
\caption{Comparison with state-of-the-art methods on THUMOS-14 and ActivityNet-1.3.}
\label{tab:sota_compare}
\resizebox{\textwidth}{!}{
\begin{tabular}{c|c|cccccc|cccc}
\toprule
\multirow{2}{*}{Method} & \multirow{2}{*}{Stage} & \multicolumn{6}{c|}{THUMOS-14} & \multicolumn{4}{c}{ActivityNet-1.3} \\
\cmidrule(lr){3-8} \cmidrule(lr){9-12}
& & 0.3 & 0.4 & 0.5 & 0.6 & 0.7 & Avg & 0.5 & 0.75 & 0.95 & Avg \\
\midrule
\multicolumn{12}{c}{\cellcolor[HTML]{EFEFEF}\textbf{Discriminative models}} \\
\midrule
TAL-Net~\cite{chao2018rethinking} & 2-stage & 53.2 & 48.5 & 42.8 & 33.8 & 20.8 & -- & 38.2 & 18.3 & 1.3 & 20.2 \\
BMN~\cite{lin2019bmn} & 2-stage & 56.0 & 47.4 & 38.8 & 29.7 & 20.5 & 38.5 & 50.1 & 34.8 & 8.3 & 33.9 \\
G-TAD~\cite{xu2020g} & 2-stage & 54.5 & 47.6 & 40.3 & 30.8 & 23.4 & 39.3 & 50.4 & 34.6 & 9.0 & 34.1 \\
RTD-Net~\cite{tan2021relaxed} & 2-stage & 68.3 & 62.3 & 51.9 & 38.8 & 23.7 & -- & 47.2 & 30.7 & 8.6 & 30.8 \\
TCANet~\cite{qing2021temporal} & 2-stage & 60.6 & 53.2 & 44.6 & 36.8 & 26.7 & 44.3 & 52.3 & 36.7 & 6.9 & 35.5 \\
ContextLoc~\cite{zhu2021enriching} & 2-stage & 68.3 & 63.8 & 54.3 & 41.8 & 26.2 & 50.9 & 56.0 & 35.2 & 3.6 & 34.2 \\
ReAct~\cite{shi2022react} & 1-stage & 69.2 & 65.0 & 57.1 & 47.8 & 35.6 & 55.0 & 49.6 & 33.0 & 8.6 & 32.6 \\
TAGS~\cite{nag2022gsm} & 1-stage & 68.6 & 63.8 & 57.0 & 46.3 & 31.8 & 52.8 & 56.3 & 36.8 & 9.6 & 36.5 \\
TALLFormer~\cite{Tallformer} & 1-stage & 76.0 & -- & 63.2 & -- & 34.5 & 59.2 & 54.1 & 36.2 & 7.9 & 35.6 \\
ActionFormer~\cite{Zhang2022_E4DJG4I4} & 1-stage & 82.1 & 77.8 & 71.0 & 59.4 & 43.9 & 66.8 & 54.7 & 37.8 & 8.4 & 36.6 \\
TemporalMaxer~\cite{TemporalMaxer} & 1-stage & 82.8 & 78.9 & 71.8 & 60.5 & 44.7 & 67.7 & -- & -- & -- & -- \\
TFFormer~\cite{TFFormer} & 1-stage & 82.1 & 78.9 & 72.0 & 60.8 & 44.9 & 67.8 & 54.4 & 36.7 & 7.5 & 35.8 \\
TransGMC~\cite{TransGMC} & 1-stage & 82.3 & 78.8 & 71.4 & 60.0 & 45.1 & 67.5 & 54.8 & 37.6 & 8.5 & 36.7 \\
TriDet~\cite{Tridet} & 1-stage & 83.6 & 80.1 & 72.9 & 62.4 & 47.4 & 69.3 & 54.7 & 38.0 & 8.4 & 36.8 \\
CLTDR-GMG~\cite{Li2024_CLTDR} & 1-stage & 84.1 & 80.3 & 73.6 & 62.4 & 48.2 & 69.9 & 55.0 & 38.0 & 8.6 & 37.1 \\
\midrule
\multicolumn{12}{c}{\cellcolor[HTML]{EFEFEF}\textbf{Generative models}} \\
\midrule
DiffTAD baseline (10-step)~\cite{Nag2023_DiffTAD} & 2-stage & 70.0 & 66.5 & 60.6 & 47.5 & 36.9 & 56.3 & 51.0 & 32.9 & 9.0 & 32.4 \\
DiffTAD (5-step)~\cite{Nag2023_DiffTAD} & 1-stage & 73.4 & 71.5 & 69.9 & 62.8 & 58.4 & 67.2 & 55.2 & 36.8 & 8.9 & 36.0 \\
DiffTAD (10-step)~\cite{Nag2023_DiffTAD} & 1-stage & 74.9 & 72.8 & 71.2 & 62.9 & 58.5 & 68.0 & 56.1 & 36.9 & 9.0 & 36.1 \\
\midrule
MDVLM-TAL-8B (ours) & 1-stage & \textbf{84.6} & \textbf{80.8} & \textbf{74.4} & \textbf{63.5} & \textbf{60.1} & \textbf{72.7} & \textbf{58.2} & \textbf{38.8} & \textbf{9.8} & \textbf{37.9} \\
\bottomrule
\end{tabular}
}
\end{table*}

\subsection{Comparison with State-of-the-Art}


Table~\ref{tab:sota_compare} shows that MDVLM-TAL improves the strongest prior results on both closed-set benchmarks. On THUMOS-14, MDVLM-TAL reaches the highest mAP at every reported tIoU threshold, with the largest gains at stricter thresholds such as 0.7. On ActivityNet-1.3, it also improves the official average mAP and the high-overlap metrics. These gains are important because high-threshold mAP measures boundary quality rather than only coarse action recognition. The results support the central claim that temporal boundaries benefit from remaining editable during sequence generation, rather than being produced once by proposal refinement or left to right decoding.


Table~\ref{tab:rtl_compare} evaluates the query-conditioned reasoning setting. ActivityNet-RTL measures localization quality with mIoU and precision-at-threshold, and measures explanation quality with the LITA ``Score''~\cite{Huang2024_RP67QJDE}. Some general-purpose Video-LLMs in the source benchmark do not reliably emit timestamps, so their comparison is score-only. We keep this reporting convention and mark $P@0.75$ and $P@0.95$ as unavailable when a baseline does not report them. We additionally evaluate Qwen3.5-9B as a stronger autoregressive baseline, using the same SigLIP-SO400M visual encoder, multimodal projector, token compressor, training data, and training settings. MDVLM-TAL achieves the highest reported localization and explanation scores while using a single 8B language backbone.

\begin{table}[t]
\caption{Results on ActivityNet-RTL under the LITA evaluation protocol~\cite{Huang2024_RP67QJDE}. Qwen3.5-9B is an additional autoregressive (AR) baseline trained on the same data with the same visual encoder, projector, and token compressor. Unreported metrics are marked ``--''.}
\vspace{3mm}
\label{tab:rtl_compare}
\centering
\begin{tabular}{lcccccc}
\toprule
Model & Size & mIOU & P@0.5 & P@0.75 & P@0.95 & Score \\\midrule
Video-LLaMA-v2~\cite{Zhang2023_VideoLLaMA} & 13B & -- & -- & -- & -- & 32.1 \\
Video-ChatGPT~\cite{Maaz2023VideoChatGPT} & 13B & -- & -- & -- & -- & 38.8 \\
LITA-7B~\cite{Huang2024_RP67QJDE} & 7B & 24.1 & 21.2 & -- & -- & 44.0 \\
LITA-13B~\cite{Huang2024_RP67QJDE} & 13B & 28.6 & 25.9 & -- & -- & 46.3 \\
Qwen3.5-9B (AR) & 9B & 30.09 & 26.59 & -- & -- & 48.3 \\
\midrule
MDVLM-TAL (ours) & 8B & \textbf{31.24} & \textbf{27.12} & \textbf{21.3} & \textbf{6.4} & \textbf{48.9} \\
\bottomrule
\end{tabular}
\end{table}

Table~\ref{tab:rtl_compare} shows that Qwen3.5-9B improves over LITA-13B. MDVLM-TAL further exceeds Qwen3.5-9B by 1.15 percentage points in mIoU, 0.53 points in $P@0.5$, and 0.6 in explanation Score. The advantage persists against this stronger autoregressive baseline under a shared visual frontend and task training setup. Since the language backbones differ, we next isolate the proposed objectives within the same masked diffusion model.

\subsection{Ablation Study}
\noindent \textbf{Cross-dataset component ablations.}
Table~\ref{tab:ablation_components} evaluates all three benchmarks with the same LLaDA-8B backbone, visual frontend, and training data. Base MDVLM retains LoRA adaptation but uses uniform fine-tuning without Planned Training or the Step-Level IoU Reward. The reversed plan recovers time tokens earlier; rewarding every step removes the both-boundary gate. Full fine-tuning replaces LoRA with full-parameter language-backbone updates. Other variants remove one component from the full model.

\begin{table}[t]
\centering
\caption{Component ablations on three benchmarks with the same 8B masked diffusion backbone and training data. ``w/o'' removes a component from the full model. All metrics are percentages except Score.}
\label{tab:ablation_components}
\small
\setlength{\tabcolsep}{3pt}
\renewcommand{\arraystretch}{1.12}
\resizebox{\linewidth}{!}{%
\begin{tabular}{@{}lrrrrrrrrrrrrr@{}}
\toprule
\multirow{2}{*}{Setting} & \multicolumn{6}{c}{THUMOS-14 (mAP)} & \multicolumn{4}{c}{ActivityNet-1.3 (mAP)} & \multicolumn{3}{c}{ActivityNet-RTL} \\
\cmidrule(lr){2-7}\cmidrule(lr){8-11}\cmidrule(lr){12-14}
& 0.3 & 0.4 & 0.5 & 0.6 & 0.7 & Avg & 0.5 & 0.75 & 0.95 & Avg & mIoU & $P@0.5$ & Score \\
\midrule
\textbf{MDVLM-TAL (ours)} & \textbf{84.6} & \textbf{80.8} & \textbf{74.4} & \textbf{63.5} & \textbf{60.1} & \textbf{72.7} & \textbf{58.2} & \textbf{38.8} & \textbf{9.8} & \textbf{37.9} & \textbf{31.24} & \textbf{27.12} & \textbf{48.9} \\
\midrule
w/o Planned Training & 83.1 & 79.0 & 72.8 & 62.3 & 57.6 & 71.0 & 57.0 & 37.8 & 9.3 & 36.9 & 28.21 & 23.86 & 46.2 \\
w/o planned masking & 83.9 & 79.8 & 73.4 & 62.6 & 58.8 & 71.7 & 57.5 & 38.1 & 9.4 & 37.2 & 29.35 & 25.14 & 47.0 \\
w/o step weighting & 84.2 & 80.2 & 73.8 & 63.0 & 59.3 & 72.1 & 57.7 & 38.3 & 9.5 & 37.3 & 30.31 & 26.02 & 48.4 \\
w/ reversed plan & 82.8 & 78.7 & 72.3 & 61.8 & 57.0 & 70.5 & 56.7 & 37.5 & 9.1 & 36.7 & 27.86 & 23.41 & 45.9 \\
\midrule
w/o IoU reward & 83.8 & 79.6 & 73.2 & 62.4 & 58.3 & 71.5 & 57.4 & 38.1 & 9.4 & 37.2 & 28.74 & 24.52 & 46.5 \\
w/ reward at every step & 84.1 & 80.1 & 73.7 & 62.9 & 59.2 & 72.0 & 57.7 & 38.2 & 9.5 & 37.3 & 30.12 & 25.86 & 48.2 \\
w/ boundary-only CE & 84.0 & 79.9 & 73.5 & 62.7 & 58.9 & 71.8 & 57.6 & 38.2 & 9.4 & 37.3 & 29.68 & 25.43 & 47.6 \\
\midrule
Full fine-tuning & 84.3 & 80.3 & 73.4 & 63.2 & 59.4 & 72.2 & 57.8 & 38.5 & 9.6 & 37.4 & 30.58 & 26.03 & 47.3 \\
Base MDVLM (w/o both) & 82.4 & 78.2 & 71.6 & 61.2 & 55.9 & 69.9 & 56.4 & 37.2 & 8.9 & 36.3 & 26.94 & 22.42 & 45.2 \\
\bottomrule
\end{tabular}
}
\end{table}

\noindent \textbf{Contributions of Planned Training.}
Removing Planned Training lowers average mAP by 1.7 points on THUMOS-14 and 1.0 point on ActivityNet-1.3, and RTL mIoU by 3.03 points. Both planned masking and step weighting contribute, with larger drops from removing masking on all three datasets. Reversing the plan is weaker still, yielding 70.5 and 36.7 average mAP and 27.86 RTL mIoU. The consistent degradation supports recovering semantic context before temporal boundaries across both closed-set detection and query-conditioned reasoning.

\noindent \textbf{Overlap supervision and boundary gating.}
Removing the Step-Level IoU Reward lowers average mAP by 1.2 and 0.7 points on THUMOS-14 and ActivityNet-1.3, and RTL mIoU by 2.5 points. Rewarding every step yields 30.12 RTL mIoU versus 31.24 for the full model, supporting overlap feedback after both boundaries are available. Boundary-only cross-entropy also underperforms across datasets, reaching 29.68 RTL mIoU and 47.6 Score. Full-sequence supervision thus benefits both localization and explanation quality.

\begin{figure}[t]
    \centering
    \caption{Illustrative single-sample denoising trajectories for selected component variants on ActivityNet-RTL. Temporal IoU is shown after a valid boundary pair first appears. Full MDVLM-TAL starts later and reaches higher final overlap than the weaker settings.}
    \includegraphics[width=\linewidth]{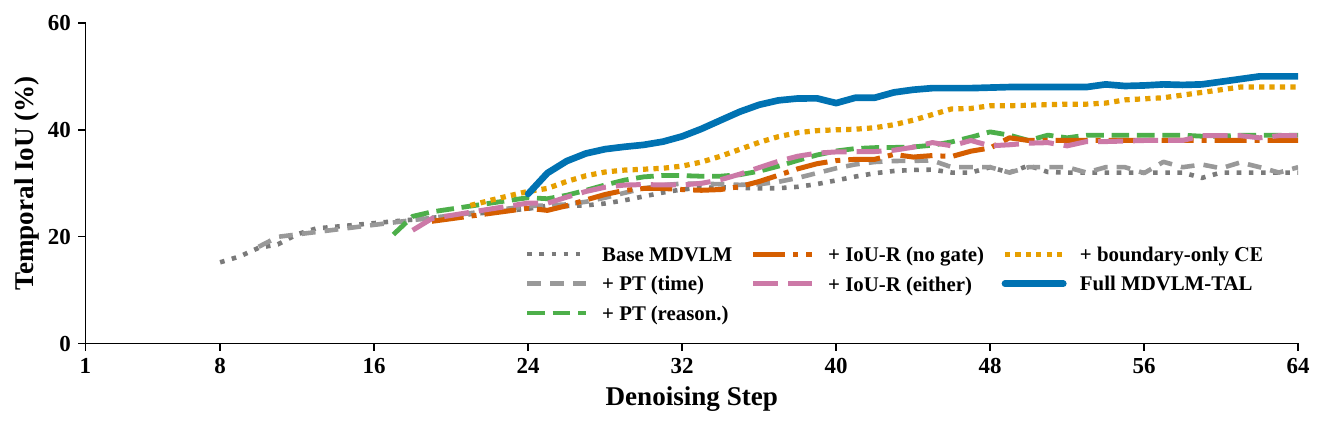}
    \label{fig:ablation_curve}
    \par\medskip
    \caption{Dataset-level mean reveal and boundary-stabilization steps, with and without Planned Training (PT). Progress $s/S$ counts forward through $S=64$ decoding updates; larger values mean later recovery. Semantic tokens emerge earlier than time tokens on all three benchmarks, and PT increases this separation.}
    \includegraphics[width=\linewidth]{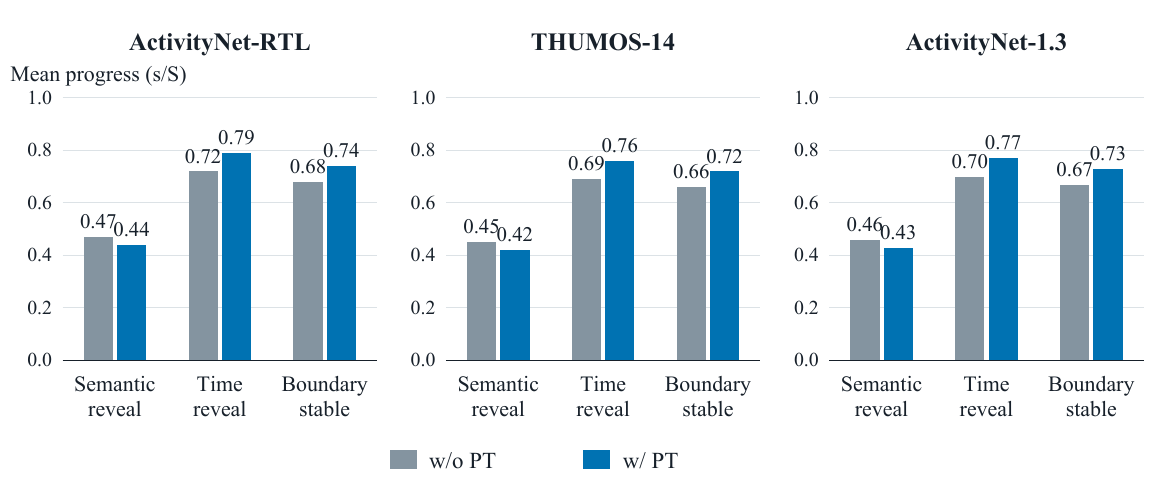}
    \label{fig:dataset_recovery}
\end{figure}

\noindent \textbf{Task-specific objectives and adaptation.}
The full model improves over Base MDVLM by 2.8 and 1.6 points in average mAP on THUMOS-14 and ActivityNet-1.3, and by 4.30 points in RTL mIoU, with LoRA and training data held fixed. Full fine-tuning obtains 72.2 and 37.4 average mAP and 30.58 RTL mIoU, below the default adaptation. Figure~\ref{fig:ablation_curve} illustrates selected variants on one sample; Figure~\ref{fig:dataset_recovery} adds dataset-level evidence for the recovery order.

\noindent \textbf{Dataset-level recovery order.}
We aggregate token reveal steps over each benchmark's evaluation examples and define boundary stabilization as the last update at which the boundary prediction changes. In Figure~\ref{fig:dataset_recovery}, $s/S$ counts decoding progress forward, unlike the decreasing corruption index $t$ in Section~\ref{sec:method}.

Without PT, semantic tokens emerge at $0.45$--$0.47S$ on average, preceding time tokens at $0.69$--$0.72S$. PT advances semantic reveal to $0.42$--$0.44S$ and delays time-token reveal to $0.76$--$0.79S$, increasing the within-dataset gap from $0.24$--$0.25S$ to $0.34$--$0.35S$. Boundary stabilization likewise shifts later, from $0.66$--$0.68S$ to $0.72$--$0.74S$. Together with the reversed-plan ablation, these results support semantic recovery before boundary determination beyond the single-sample trajectory. The means describe a dataset-level tendency, rather than an identical order for every example.

\begin{table*}[t]
  \caption{Loss-weight ablation on THUMOS-14 and ActivityNet-1.3 for the 8B MDVLM-TAL model.}
\label{tab:ablation_weights}
\vspace{3mm}
\centering
\begin{tabular}{ccc|ccccc|ccc}
\toprule
\multirow{2}{*}{$\lambda_{\text{plan}}$} & \multirow{2}{*}{$\lambda_{\text{iou}}$} & \multirow{2}{*}{$\lambda_{\text{ce}}$} & \multicolumn{5}{c|}{THUMOS-14} & \multicolumn{3}{c}{ActivityNet-1.3} \\
\cmidrule(lr){4-8}\cmidrule(lr){9-11}
& & & 0.3 & 0.4 & 0.5 & 0.6 & 0.7 & 0.5 & 0.75 & 0.95 \\
\midrule
1.0 & 0.25 & 0.2 & 83.9 & 80.1 & 73.6 & 62.8 & 59.4 & 57.5 & 38.1 & 9.4 \\
1.0 & 0.5 & 0.1 & 84.2 & 80.4 & 73.9 & 63.1 & 59.7 & 57.8 & 38.4 & 9.6 \\
\textbf{1.0} & \textbf{0.5} & \textbf{0.2} & \textbf{84.6} & \textbf{80.8} & \textbf{74.4} & \textbf{63.5} & \textbf{60.1} & \textbf{58.2} & \textbf{38.8} & \textbf{9.8} \\
1.0 & 1.0 & 0.2 & 84.0 & 80.2 & 73.7 & 62.9 & 59.5 & 57.6 & 38.2 & 9.5 \\
0.5 & 0.5 & 0.2 & 83.7 & 79.9 & 73.4 & 62.6 & 59.2 & 57.3 & 37.9 & 9.3 \\
\bottomrule
\end{tabular}
\end{table*}
\noindent \textbf{Loss Weights.}
Table~\ref{tab:ablation_weights} reports a local sweep on THUMOS-14 and ActivityNet-1.3 while keeping the 8B architecture and both-boundary gate fixed. The default setting $(1.0,0.5,0.2)$ performs best across the two benchmark families. Reducing $\lambda_{\text{plan}}$ weakens the alignment between training corruption and boundary recovery. Increasing $\lambda_{\text{iou}}$ overemphasizes local overlap optimization, and reducing $\lambda_{\text{ce}}$ weakens the dense sequence supervision that couples labels with boundaries. The same setting also gives the strongest high-tIoU results, which indicates that the selected balance improves precise boundary placement rather than only coarse detection.

\section{Conclusion}

We introduced MDVLM-TAL, a masked diffusion vision-language model for temporal action localization that jointly denoises semantic reasoning tokens and temporal boundary tokens. Instead of predicting timestamps in a fixed left to right order, MDVLM-TAL keeps the full response editable throughout reverse denoising, so semantic evidence and temporal evidence can refine each other with bidirectional context. This reframes temporal localization as a structured sequence-generation problem. In this view, localization quality depends not only on the final boundary tokens but also on how those tokens emerge during denoising.

From this perspective, we proposed some designs: a planned objective that keeps time tokens masked longer than semantic tokens and emphasizes lower-noise denoising steps, a step-level IoU reward that provides overlap-aware supervision during denoising, and a standard cross-entropy reconstruction term that stabilizes sequence generation. Our ablation results show that these components contribute in distinct but consistent ways: planned masking improves the denoising schedule, IoU-based reward improves boundary accuracy, and dense sequence supervision improves the coupling between semantics and localization. More broadly, MDVLM-TAL points toward a unified class of video-language models that treats localization, captioning, and reasoning as parts of the same vision-language generation process rather than as isolated modules.

\clearpage
\bibliography{references}

@inproceedings{Huang2024_RP67QJDE,
  title = {LITA: Language Instructed Temporal-Localization Assistant},
  author = {Huang, De-An and Liao, Shijia and Radhakrishnan, Subhashree and Yin, Hongxu and Molchanov, Pavlo and Yu, Zhiding and Kautz, Jan},
  booktitle = {ECCV},
  pages = {202--218},
  publisher = {Springer Nature Switzerland},
  doi = {10.1007/978-3-031-73039-9_12},
  year = {2024}
}

@inproceedings{Zeng2024_W77Z2KFV,
  title = {UniMD: Towards Unifying Moment Retrieval and Temporal Action Detection},
  author = {Zeng, Yingsen and Zhong, Yujie and Feng, Chengjian and Ma, Lin},
  booktitle = {ECCV},
  pages = {286--304},
  publisher = {Springer Nature Switzerland},
  doi = {10.1007/978-3-031-72952-2_17},
  year = {2024}
}

@inproceedings{zhai2023siglip,
  title={Sigmoid loss for language image pre-training},
  author={Zhai, Xiaohua and Mustafa, Basil and Kolesnikov, Alexander and Beyer, Lucas},
  booktitle={ICCV},
  pages={11975--11986},
  year={2023}
}

@inproceedings{wang2024internvideo2,
  title={Internvideo2: Scaling foundation models for multimodal video understanding},
  author={Wang, Yi and Li, Kunchang and Li, Xinhao and Yu, Jiashuo and He, Yinan and Chen, Guo and Pei, Baoqi and Zheng, Rongkun and Wang, Zun and Shi, Yansong and others},
  booktitle={ECCV},
  pages={396--416},
  year={2024},
  organization={Springer}
}

@article{wang2022internvideo,
  title={Internvideo: General video foundation models via generative and discriminative learning},
  author={Wang, Yi and Li, Kunchang and Li, Yizhuo and He, Yinan and Huang, Bingkun and Zhao, Zhiyu and Zhang, Hongjie and Xu, Jilan and Liu, Yi and Wang, Zun and others},
  doi={10.48550/arXiv.2212.03191},
  year={2022}
}

@article{wang2025internvideo2,
  title={Internvideo2.5: Empowering video mllms with long and rich context modeling},
  author={Wang, Yi and Li, Xinhao and Yan, Ziang and He, Yinan and Yu, Jiashuo and Zeng, Xiangyu and Wang, Chenting and Ma, Changlian and Huang, Haian and Gao, Jianfei and others},
  doi={10.48550/arXiv.2501.12386},
  year={2025}
}

@article{bai2025qwen3,
  title={Qwen3-vl technical report},
  author={Bai, Shuai and Cai, Yuxuan and Chen, Ruizhe and Chen, Keqin and Chen, Xionghui and Cheng, Zesen and Deng, Lianghao and Ding, Wei and Gao, Chang and Ge, Chunjiang and others},
  doi={10.48550/arXiv.2511.21631},
  year={2025}
}

@inproceedings{Heilbron2015_CX49Q6U2,
  title = {ActivityNet: A Large-Scale Video Benchmark for Human Activity Understanding},
  author = {Heilbron, Fabian and Escorcia, Victor and Ghanem, Bernard and Niebles, Juan Carlos},
  booktitle = {CVPR},
  year = {2015}
}

@article{nie2025lldm,
  title={Large language diffusion models},
  author={Nie, Shen and Zhu, Fengqi and You, Zebin and Zhang, Xiaolu and Ou, Jingyang and Hu, Jun and Zhou, Jun and Lin, Yankai and Wen, Ji-Rong and Li, Chongxuan},
  doi={10.48550/arXiv.2502.09992},
  year={2025}
}

@inproceedings{li2025lavida,
  title={Lavida: A large diffusion model for vision-language understanding},
  author={Li, Shufan and Kallidromitis, Konstantinos and Bansal, Hritik and Gokul, Akash and Kato, Yusuke and Kozuka, Kazuki and Kuen, Jason and Lin, Zhe and Chang, Kai-Wei and Grover, Aditya},
  booktitle={NeurIPS},
  year={2025}
}

@article{hu2022lora,
  title={Lora: Low-rank adaptation of large language models.},
  author={Hu, Edward J and Shen, Yelong and Wallis, Phillip and Allen-Zhu, Zeyuan and Li, Yuanzhi and Wang, Shean and Wang, Liang and Chen, Weizhu and others},
  journal={ICLR},
  volume={1},
  number={2},
  pages={3},
  year={2022}
}

@inproceedings{Zhang2022_E4DJG4I4,
  title={Actionformer: Localizing moments of actions with transformers},
  author={Zhang, Chen-Lin and Wu, Jianxin and Li, Yin},
  booktitle={ECCV},
  pages={492--510},
  year={2022},
  organization={Springer}
}

@inproceedings{Nag2023_DiffTAD,
  title={Difftad: Temporal action detection with proposal denoising diffusion},
  author={Nag, Sauradip and Zhu, Xiatian and Deng, Jiankang and Song, Yi-Zhe and Xiang, Tao},
  booktitle={CVPR},
  year={2023}
}

@inproceedings{Li2024_CLTDR,
  title={Temporal action localization with cross layer task decoupling and refinement},
  author={Li, Qiang and Liu, Di and Kong, Jun and Li, Sen and Xu, Hui and Wang, Jianzhong},
  booktitle={AAAI},
  pages={4878--4886},
  year={2025}
}

@inproceedings{chao2018rethinking,
  title = {Rethinking the Faster R-CNN Architecture for Temporal Action Localization},
  author = {Chao, Yu-Wei and Vijayanarasimhan, Sudheendra and Seybold, Bryan and Ross, David A. and Deng, Jia and Sukthankar, Rahul},
  booktitle = {CVPR},
  year = {2018}
}

@inproceedings{lin2019bmn,
  title = {BMN: Boundary-Matching Network for Temporal Action Proposal Generation},
  author = {Lin, Tianwei and Liu, Xiao and Li, Xin and Ding, Errui and Wen, Shilei},
  booktitle = {ICCV},
  pages = {3889--3898},
  year = {2019}
}

@inproceedings{xu2020g,
  title = {G-TAD: Sub-Graph Localization for Temporal Action Detection},
  author = {Xu, Mengmeng and Zhao, Chen and Rojas, David S. and Thabet, Ali and Ghanem, Bernard},
  booktitle = {CVPR},
  year = {2020}
}

@inproceedings{tan2021relaxed,
  title = {Relaxed Transformer Decoders for Direct Action Proposal Generation},
  author = {Tan, Jing and Tang, Jiaqi and Wang, Limin and Wu, Gangshan},
  booktitle = {ICCV},
  year = {2021}
}

@inproceedings{qing2021temporal,
  title = {Temporal Context Aggregation Network for Temporal Action Proposal Refinement},
  author = {Qing, Zhiwu and Su, Haisheng and Gan, Weihao and Wang, Dongliang and Wu, Wei and Wang, Xiang and Qiao, Yu and Yan, Junjie and Gao, Changxin and Sang, Nong},
  booktitle = {CVPR},
  pages = {485--494},
  year = {2021}
}

@inproceedings{zhu2021enriching,
  title = {Enriching Local and Global Contexts for Temporal Action Localization},
  author = {Zhu, Zixin and Tang, Wei and Wang, Le and Zheng, Nanning and Hua, Gang},
  booktitle = {ICCV},
  pages = {13516--13525},
  year = {2021}
}

@inproceedings{shi2022react,
  title = {ReAct: Temporal Action Detection with Relational Queries},
  author = {Shi, Dingfeng and Zhong, Yujie and Cao, Qiong and Zhang, Jing and Ma, Lin and Li, Jia and Tao, Dacheng},
  booktitle = {ECCV},
  year = {2022}
}

@inproceedings{nag2022gsm,
  title = {Proposal-Free Temporal Action Detection via Global Segmentation Mask Learning},
  author = {Nag, Sauradip and Zhu, Xiatian and Song, Yi-Zhe and Xiang, Tao},
  booktitle = {ECCV},
  year = {2022}
}

@inproceedings{Tallformer,
  title = {TallFormer: Temporal Action Localization with a Long-Memory Transformer},
  author = {Cheng, Feng and Bertasius, Gedas},
  booktitle = {ECCV},
  pages = {503--521},
  year = {2022}
}

@article{TemporalMaxer,
  title={Temporalmaxer: Maximize temporal context with only max pooling for temporal action localization},
  author={Tang, Tuan N and Kim, Kwonyoung and Sohn, Kwanghoon},
  doi={10.48550/arXiv.2303.09055},
  year={2023}
}

@article{TFFormer,
  title = {Cross Time-Frequency Transformer for Temporal Action Localization},
  author = {Yang, Jin and Wei, Ping and Zheng, Nanning},
  journal = {IEEE Transactions on Circuits and Systems for Video Technology},
  year = {2024},
  volume = {34},
  number = {6},
  pages = {4625--4638},
  doi = {10.1109/TCSVT.2023.3326692}
}

@article{TransGMC,
  title = {Gated Multi-Scale Transformer for Temporal Action Localization},
  author = {Yang, Jin and Wei, Ping and Ren, Ziyang and Zheng, Nanning},
  journal = {IEEE Transactions on Multimedia},
  year = {2024},
  volume = {26},
  pages = {5705--5717},
  doi = {10.1109/TMM.2023.3338082}
}

@inproceedings{Tridet,
  title = {TriDet: Temporal Action Detection with Relative Boundary Modeling},
  author = {Shi, Dingfeng and Zhong, Yujie and Cao, Qiong and Ma, Lin and Li, Jia and Tao, Dacheng},
  booktitle = {CVPR},
  pages = {18857--18866},
  year = {2023}
}

@article{Guo2024_ZZVDVZ6M,
  title = {VTG-LLM: Integrating Timestamp Knowledge into Video LLMs for Enhanced Video Temporal Grounding},
  author = {Guo, Yongxin and Liu, Jingyu and Li, Mingda and Cheng, Dingxin and Tang, Xiaoying and Sui, Dianbo and Liu, Qingbin and Chen, Xi and Zhao, Kevin},
  journal = {AAAI},
  volume = {39},
  number = {3},
  pages = {3302--3310},
  publisher = {Association for the Advancement of Artificial Intelligence (AAAI)},
  doi = {10.1609/aaai.v39i3.32341},
  year = {2025}
}

@inproceedings{Huang2024_E99RJNWL,
  title={Vtimellm: Empower llm to grasp video moments},
  author={Huang, Bin and Wang, Xin and Chen, Hong and Song, Zihan and Zhu, Wenwu},
  booktitle={CVPR},
  pages={14271--14280},
  year={2024}
}

@inproceedings{Zhang2024_XRTWDU53,
  title = {Weakly Supervised Temporal Action Localization via Dual-Prior Collaborative Learning Guided by Multimodal Large Language Models},
  author = {Zhang, Quan and Fang, Jinwei and Yuan, Rui and Tang, Xi and Qi, Yuxin and Zhang, Ke and Yuan, Chun},
  booktitle = {CVPR},
  pages = {24139--24148},
  publisher = {IEEE},
  doi = {10.1109/CVPR52734.2025.02248},
  year = {2025}
}

@article{Xu2024_I9Z8BZ8F,
  title={Slowfast-llava: A strong training-free baseline for video large language models},
  author={Xu, Mingze and Gao, Mingfei and Gan, Zhe and Chen, Hong-You and Lai, Zhengfeng and Gang, Haiming and Kang, Kai and Dehghan, Afshin},
  doi={10.48550/arXiv.2407.15841},
  year={2024}
}

@article{Xu2024_FM5VVXFF,
  title={Pllava: Parameter-free llava extension from images to videos for video dense captioning},
  author={Xu, Lin and Zhao, Yilin and Zhou, Daquan and Lin, Zhijie and Ng, See Kiong and Feng, Jiashi},
  doi={10.48550/arXiv.2404.16994},
  year={2024}
}

@misc{Liu2023_LLaVA,
  title = {Visual Instruction Tuning},
  doi = {10.48550/arXiv.2304.08485},
  author = {Liu, Haotian and Li, Chunyuan and Wu, Qingyang and Lee, Yong Jae},
  year = {2023}
}

@inproceedings{Zhang2023_VideoLLaMA,
  title = {Video-LLaMA: An Instruction-tuned Audio-Visual Language Model for Video Understanding},
  author = {Zhang, Hang and Li, Xin and Bing, Lidong},
  booktitle = {EMNLP Demo},
  pages = {543--553},
  publisher = {Association for Computational Linguistics},
  doi = {10.18653/v1/2023.emnlp-demo.49},
  year = {2023}
}

@inproceedings{Maaz2023VideoChatGPT,
  title={Video-chatgpt: Towards detailed video understanding via large vision and language models},
  author={Maaz, Muhammad and Rasheed, Hanoona and Khan, Salman and Khan, Fahad},
  booktitle={ACL},
  pages={12585--12602},
  year={2024}
}

@article{Li2023_VideoChat,
  title = {VideoChat: chat-centric video understanding},
  author = {Li, Kunchang and He, Yinan and Wang, Yi and Li, Yizhuo and Wang, Wenhai and Luo, Ping and Wang, Yali and Wang, Limin and Qiao, Yu},
  journal = {Science China Information Sciences},
  volume = {68},
  number = {10},
  publisher = {Springer Science and Business Media LLC},
  doi = {10.1007/s11432-024-4321-9},
  year = {2025}
}

@misc{Yu2023_SeViLA,
  title = {SeViLA: Self-Chained Image-Language Model for Video Localization and Question Answering},
  doi = {10.48550/arXiv.2305.06988},
  author = {Yu, Shoubin and Cho, Jaemin and Yadav, Prateek and Bansal, Mohit},
  year = {2023}
}

@inproceedings{JX7MI83H,
  title={Test-time zero-shot temporal action localization},
  author={Liberatori, Benedetta and Conti, Alessandro and Rota, Paolo and Wang, Yiming and Ricci, Elisa},
  booktitle={CVPR},
  pages={18720--18729},
  year={2024}
}

@inproceedings{ZEUUYWHT,
  title={Improving weakly supervised temporal action localization by bridging train-test gap in pseudo labels},
  author={Zhou, Jingqiu and Huang, Linjiang and Wang, Liang and Liu, Si and Li, Hongsheng},
  booktitle={CVPR},
  pages={23003--23012},
  year={2023}
}

@article{UBV766ZB,
  title = {Adaptive Two-Stream Consensus Network for Weakly-Supervised Temporal Action Localization},
  author = {Zhai, Yuanhao and Wang, Le and Tang, Wei and Zhang, Qilin and Zheng, Nanning and Doermann, David and Yuan, Junsong and Hua, Gang},
  journal = {IEEE Transactions on Pattern Analysis and Machine Intelligence},
  volume = {45},
  number = {4},
  pages = {4136--4151},
  doi = {10.1109/TPAMI.2022.3189662},
  year = {2023}
}

@article{9NG3983D,
  title = {Weakly-supervised Temporal Action Localization by Uncertainty Modeling},
  author = {Lee, Pilhyeon and Wang, Jinglu and Lu, Yan and Byun, Hyeran},
  journal = {AAAI},
  volume = {35},
  number = {3},
  pages = {1854--1862},
  year = {2021}
}

@inproceedings{XTT9IIB7,
  title={Weakly supervised temporal action localization via representative snippet knowledge propagation},
  author={Huang, Linjiang and Wang, Liang and Li, Hongsheng},
  booktitle={CVPR},
  pages={3272--3281},
  year={2022}
}

@inproceedings{DTVJ7N9S,
  title = {Proposal-Based Multiple Instance Learning for Weakly-Supervised Temporal Action Localization},
  author = {Ren, Huan and Yang, Wenfei and Zhang, Tianzhu and Zhang, Yongdong},
  booktitle={CVPR},
  pages = {2394--2404},
  year = {2023}
}

@inproceedings{JXFQ498U,
  title={Online temporal action localization with memory-augmented transformer},
  author={Song, Youngkil and Kim, Dongkeun and Cho, Minsu and Kwak, Suha},
  booktitle={ECCV},
  pages={74--91},
  year={2024},
  organization={Springer}
}

@inproceedings{5F2AFWKI,
  title={Hat: History-augmented anchor transformer for online temporal action localization},
  author={Reza, Sakib and Zhang, Yuexi and Moghaddam, Mohsen and Camps, Octavia},
  booktitle={ECCV},
  pages={205--222},
  year={2024},
  organization={Springer}
}

@inproceedings{V796WXUR,
  title = {OpenTAD: A Unified Framework and Comprehensive Study of Temporal Action Detection},
  author = {Liu, Shuming and Zhao, Chen and Zohra, Fatimah and Soldan, Mattia and Pardo, Alejandro and Xu, Mengmeng and Alssum, Lama and Ramazanova, Merey and Alc{\'a}zar, Juan Le{\'o}n and Cioppa, Anthony and Giancola, Silvio and Hinojosa, Carlos and Ghanem, Bernard},
  booktitle = {CVPR Workshops},
  pages = {2616--2626},
  publisher = {IEEE},
  doi = {10.1109/CVPRW67362.2025.00247},
  year = {2025}
}

@inproceedings{U84863CG,
  title = {Dual DETRs for Multi-Label Temporal Action Detection},
  author = {Zhu, Yuhan and Zhang, Guozhen and Tan, Jing and Wu, Gangshan and Wang, Limin},
  booktitle = {CVPR},
  pages = {18559--18569},
  publisher = {IEEE},
  doi = {10.1109/CVPR52733.2024.01756},
  year = {2024}
}

@article{CV922R8W,
  title = {Prediction-Feedback DETR for Temporal Action Detection},
  author = {Kim, Jihwan and Lee, Miso and Cho, Cheol-Ho and Lee, Jihyun and Heo, Jae-Pil},
  journal = {AAAI},
  volume = {39},
  number = {4},
  pages = {4266--4274},
  publisher = {Association for the Advancement of Artificial Intelligence (AAAI)},
  doi = {10.1609/aaai.v39i4.32448},
  year = {2025}
}

@inproceedings{H8Q99P63,
  title={RefineTAD: learning proposal-free refinement for temporal action detection},
  author={Feng, Yue and Zhang, Zhengye and Quan, Rong and Wang, Limin and Qin, Jie},
  booktitle={ACM MM},
  pages={135--143},
  year={2023}
}

@article{FX979ZEP,
  title = {End-to-End Temporal Action Detection With Transformer},
  author = {Liu, Xiaolong and Wang, Qimeng and Hu, Yao and Tang, Xu and Zhang, Shiwei and Bai, Song and Bai, Xiang},
  journal = {IEEE Transactions on Image Processing},
  volume = {31},
  pages = {5427--5441},
  doi = {10.1109/TIP.2022.3195321},
  year = {2022}
}

@inproceedings{37QXWBMM,
  title={Zero-shot temporal action detection via vision-language prompting},
  author={Nag, Sauradip and Zhu, Xiatian and Song, Yi-Zhe and Xiang, Tao},
  booktitle={ECCV},
  pages={681--697},
  year={2022},
  organization={Springer}
}

@inproceedings{6SZNS42I,
  title = {ZEETAD: Adapting Pretrained Vision-Language Model for Zero-Shot End-to-End Temporal Action Detection},
  author = {Phan, Thinh and Vo, Khoa and Le, Duy and Doretto, Gianfranco and Adjeroh, Donald and Le, Ngan},
  booktitle={WACV},
  pages = {7031--7040},
  year = {2024}
}

@misc{7QIS3HW9,
  title = {Boundary-Denoising for Video Activity Localization},
  author = {Xu, Mengmeng and Soldan, Mattia and Gao, Jialin and Liu, Shuming and P{\'e}rez-R{\'u}a, Juan-Manuel and Ghanem, Bernard},
  doi = {10.48550/arXiv.2304.02934},
  year = {2023}
}

@inproceedings{peng2024harnessing,
  title={Harnessing text-to-image diffusion models for category-agnostic pose estimation},
  author={Peng, Duo and Zhang, Zhengbo and Hu, Ping and Ke, Qiuhong and Yau, David KY and Liu, Jun},
  booktitle={European Conference on Computer Vision},
  pages={342--360},
  year={2024},
  organization={Springer}
}

@inproceedings{zhang2024diff,
  title={Diff-tracker: text-to-image diffusion models are unsupervised trackers},
  author={Zhang, Zhengbo and Xu, Li and Peng, Duo and Rahmani, Hossein and Liu, Jun},
  booktitle={European Conference on Computer Vision},
  pages={319--337},
  year={2024},
  organization={Springer}
}

@article{wang2026dc,
  title={DC-Motion: Decoupling Semantics and Details via Discrete-Continuous Tokens for Human Motion Generation},
  author={Wang, Hequan and Zhang, Jiaxu and Zhang, Zhengbo and Tu, Zhigang},
  journal={arXiv preprint arXiv:2606.14721},
  year={2026}
}

@inproceedings{zhao2023unleashing,
  title={Unleashing text-to-image diffusion models for visual perception},
  author={Zhao, Wenliang and Rao, Yongming and Liu, Zuyan and Liu, Benlin and Zhou, Jie and Lu, Jiwen},
  booktitle={2023 IEEE/CVF International Conference on Computer Vision (ICCV)},
  pages={5706--5716},
  year={2023},
  organization={IEEE}
}

@inproceedings{kondapaneni2024text,
  title={Text-image alignment for diffusion-based perception},
  author={Kondapaneni, Neehar and Marks, Markus and Knott, Manuel and Guimaraes, Rog{\'e}rio and Perona, Pietro},
  booktitle={2024 IEEE/CVF Conference on Computer Vision and Pattern Recognition (CVPR)},
  pages={13883--13893},
  year={2024},
  organization={IEEE}
}
\bibliographystyle{abbrvnat}

\end{document}